\documentclass[journal]{IEEEtran}
\ifCLASSINFOpdf
\else
\fi
\ifCLASSOPTIONcompsoc
 \usepackage[caption=false,font=normalsize,labelfont=sf,textfont=sf]{subfig}
\else
 \usepackage[caption=false,font=footnotesize]{subfig}
\fi
\usepackage{graphicx}  
\usepackage{float}  
\usepackage{siunitx}
\usepackage{subfig}
\usepackage{makecell}
\usepackage{multirow}
\usepackage{algorithm}
\usepackage{algorithmic}

\usepackage{colortbl}
\usepackage{cite}
\usepackage{threeparttable}
\usepackage{amsfonts,amssymb} 
\usepackage{amsmath} 
\usepackage{amssymb}  
\DeclareMathOperator*{\argmin}{argmin}
\usepackage{algorithm}
\usepackage{algorithmic}
\begin{document}
%
\title{Speech Corpora Divergence Based Unsupervised Data Selection for ASR}
%
%
%

\author{Changfeng~Gao,~\IEEEmembership{Student Member,~IEEE,}
        Gaofeng~Cheng,~\IEEEmembership{Member,~IEEE,}
        Pengyuan~Zhang,~\IEEEmembership{Member,~IEEE,}
        Yonghong~Yan,~\IEEEmembership{Member,~IEEE,}
\thanks{The authors are with the Key Laboratory of Speech Acoustics and Content Understanding, Institute of Acoustics, Chinese Academy of Sciences,
Beijing 100864, China. University of Chinese Academy of Sciences (e-mail: gaochangfeng@hccl.ioa.ac.cn;  chenggaofeng@hccl.ioa.ac.cn; zhangpengyuan@hccl.ioa.ac.cn; yanyonghong@hccl.ioa.ac.cn}
}

\maketitle

\begin{abstract}
Selecting application scenarios matching data is important for the automatic speech recognition (ASR) training, but it is difficult to measure the matching degree of the training corpus.
This study proposes a unsupervised target-aware data selection method based on speech corpora divergence (SCD), which can measure the similarity between two speech corpora.
We first use the self-supervised Hubert model to discretize the speech corpora into label sequence and calculate the N-gram probability distribution. 
Then we calculate the Kullback-Leibler divergence between the N-grams as the SCD.
Finally, we can choose the subset which has minimum SCD to the target corpus for annotation and training.
Compared to previous data selection method, the SCD data selection method can focus on more acoustic details and guarantee the diversity of the selected set.
We evaluate our method on different accents from Common Voice. Experiments show that the proposed SCD data selection can realize 14.8\% relative improvements to the random selection, comparable or even superior to the result of supervised selection.

\end{abstract}

\section{Introduction}

For the automatic speech recognition (ASR), speech annotation is an expensive and time-consuming work. People can easily collect enormous unsupervised corpus from websites, broadcasts, and podcasts, but only a minority of them can be annotated artificially.
Moreover, as the training data matching is crucial for the ASR system \cite{is_robust,robust-wav2vec}, it is important to select a suitable subset for annotation and training according to the target scenarios like accented\cite{google-accent}, far-field\cite{ami-pre} and children\cite{childern} ASR.
Although we can manually select the matching training corpus by accent, speaker, or channel for annotation, it is difficult to describe the corpus's similarity mathematically and select the training speech automatically.  

In general, most works\cite{es-ivector,es-gmm-u,es-gmm-u2,select-text, select-text2,es-nbest,es-gmm-s} believes that a well-designed training set should have similar distribution with the target corpus, but it is difficult to measure the speech corpus distribution.
To solve this problem, the most common method is measuring the speech distribution with the transcription\cite{select-text, select-text2, es-nbest, es-gmm-s}. \cite{select-text} uses the frequency of the word, character or phoneme to measure the transcription distribution and then samples data uniformly. And to sample unlabeled speech, \cite{es-nbest} uses a baseline ASR model to decode the N-Best hypothesis and then calculate the term frequency-inverse document frequency (tf-idf) for data selection.
As the text-level distribution is too rough to measure the acoustic difference of the speech, \cite{es-gmm-s} count the distribution according to the context-dependent HMM states to capture more acoustic details. However, the HMM states still largely depends on the speech transcription and the lexicon, it can not still measure the difference between sex, accent or other acoustic characteristics.
Besides selection by the distribution, contrastive sampling\cite{contrastive-selection,cs-gmm,accent-selection,contrastive-ds2,contrastive-new,google-ds} is another recent popular data selection method. Most of them use a universal and a target domain ASR model to score the utterances by the confidence score or the hypothesis perplexity. 
Then they will sample the utterances which has largest gap between the target and the universal score one by one. 
Using different domain ASR models can evaluate the speech from the acoustic characteristics well, however, it also tend to choose similar speech and reduce the diversity of the selected set.

In this study, we design a novel target-aware data selection method by proposing the speech corpora divergence (SCD). We use the self-supervised learning (SSL) model, Hubert\cite{hubert}, to discretize the speech and then measure the speech distribution in the discrete space. 
We count the N-gram of the discrete corpus and use the Kullback-Leibler divergence (KLD) to calculate the SCD.
Then we can select a subset from the universal unlabeled corpus by minimizing the SCD between the selected corpus and the target corpus and further use greedy search to simplify the algorithm complexity.
Compared with the previous works, the Hubert discrete labels can contain both acoustic and semantic information, so it can represent the speech distribution better than the text-related labels like word, char and HMM states. And as the SCD selection method considers the relationship between the whole selected speech and the target set, it can sample more diverse speech than the contrastive sampling.
We evaluate the SCD data selection method on different accented speech from Common Voice.
Experiments prove that our proposed method can realize 14.8\% relative improvements to the random selection and reach or even exceed the human selection result with accent labels.


\section{Related Work}

Hubert\cite{hubert} is one of the most successful SSL model which has been applied on different speech tasks. The Hubert model uses a CNN feature extractor to convert the waveform into hidden representations. Then it will mask a part of representations and use a transformer encoder to predict the discrete labels of the masked part.The discrete labels are initialed by a K-means cluster on MFCC and then are refined by the Hubert model iteratively. 
Some research work find that the labels find by Hubert can be used in different tasks. 
For example, \cite{ao22_interspeech,arunkumar22_interspeech} use these labels to pre-train the ASR decoder. They find that the Hubert discrete labels can help the decoder learn how to generate text sequences. 
\cite{Lakhotia2021OnGS, Lee2021DirectST, Polyak2021hubert_res} find that they can resynthesis the speech by combining the discrete label with some pitch and speaker information. There are also some works apply these discrete labels on emotion conversion\cite{kreuk2021textless} and NLP task\cite{kharitonov2022textless}.
These researches indicates that compared to the traditional label like word or phone, the Hubert discrete labels contain richer acoustic and semantic information. So in this paper, we will use the Hubert discrete labels to represent the continues speech and then design a data selection method according to the distribution of the Hubert labels.

\section{Method}

\begin{figure}[!t]
\centering
\includegraphics[height=2.3in]{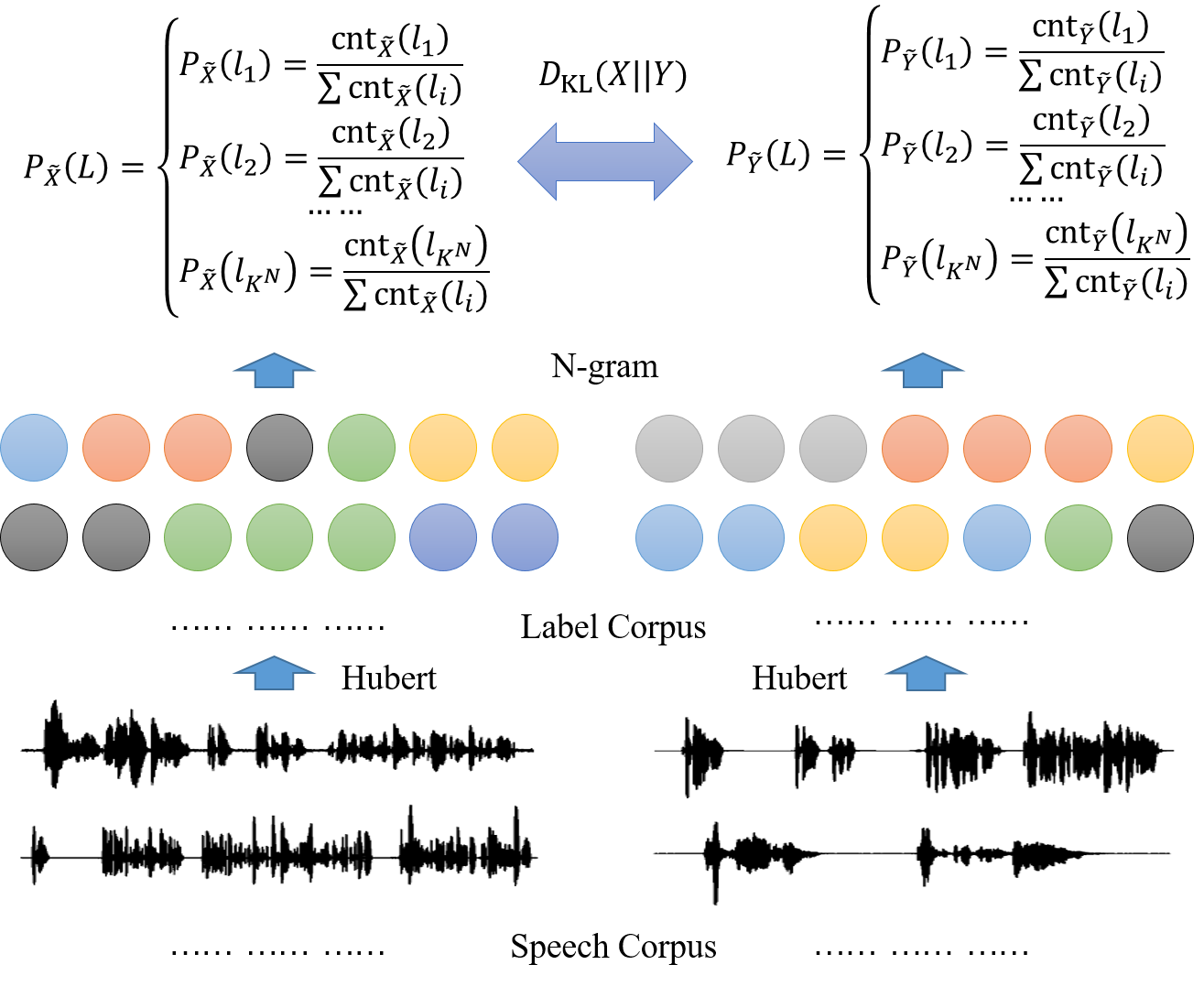}
\caption{Calculation of the SCD. We use the Hubert model to convert the speech corpora into label corpora. Then use the N-gram to measure the distribution. The SCD can be defined by the KLD between the N-grams.}
\label{fig_scd}
\end{figure}

\subsection{Speech Corpora Divergence}
In this subsection, we give the definition of the SCD, which measures the similarity between two speech corpora. 
A speech corpus $X$ can be represented by a stochastic process with probability distribution $P(X)=P(X_1X_2\dots X_t\dots)$ and each speech utterance $x_i=x_{i1}x_{i2}\dots x_{it}\dots$ is a sampling result, $i$ and $t$ stand for the utterance index and the time step.
Then we can use the KLD to measure the difference between two distribution:


\begin{equation}\label{eq_ce}
    \mathrm{SCD}(X,Y) = D_{\mathrm{KL}}(X||Y)
\end{equation} 

However, it is not easy to calculate the SCD by eq (\ref{eq_ce}), because corpora $X$ and $Y$ are in continuous space. 
Inspired by the recent SSL\cite{VQ-vae,vq-w2v,w2v,hubert,hubert2}, we use the hidden units discovery system in Hubert to discretize the speech corpora. 
For each corpus, every utterance $x_i$ will be converted into label sequence $\widetilde{x}_{i1},\widetilde{x}_{i2}\dots \widetilde{x}_{in}\dots$, and $\widetilde{x}_{in} \in \mathcal{L}$, $ \mathcal{L} = \{1,2,…,K\}$. $K$ is the clusters number of the hidden units discovery system.
After obtaining the discrete $\widetilde{X}$, we can use an N-gram model $P_{\widetilde{X}}(L)$ to represent the $P(X)$:
\begin{equation}
    P_{\widetilde{X}}(L=l_i) = \frac{\mathrm{cnt}_{\widetilde{X}}(l_i)}{\sum_{l_j} \mathrm{cnt}_{\widetilde{X}}(l_j)}
\end{equation}
where $L \in \mathcal{L}^N$, $\mathcal{L}^N$ is the N-order cartesian power of $\mathcal{L}$.
$\mathrm{cnt}_{\widetilde{X}}$ stands for the count operation in corpora $\widetilde{X}$. 
Finally the SCD can be calculated as:
\begin{equation}
\begin{aligned}
    \mathrm{SCD}(X,Y) =\sum_{l_i\in \mathcal{L}^N} P_{\widetilde{X}}(L=l_i) \log \frac{P_{\widetilde{X}}(L=l_i)}{P_{\widetilde{Y}}(L=l_i)}
\end{aligned}
\end{equation}
We conclude the calculation of the SCD in Fig.\ref{fig_scd}.

\subsection{Target-aware data selection with SCD}
With the help of the SCD, we can re-define the target-aware data selection as a combinatorial optimization problem.
Given the unlabeled universal speech corpus $U$ and the query speech set $Q$, sample a subset $S$ from $U$ with size $C$ and minimize the $\mathrm{SCD}(Q, S)$ at the same time:
\begin{equation}
\begin{aligned}
       S^* =  \argmin_S \mathrm{SCD}(Q,S), where \quad |S| = C 
\end{aligned}
\end{equation}

\begin{algorithm}[!t] 
    \caption{Target-Aware Data Selection with Speech Corpus Divergence} 
    \label{alg:data-selection} 
    \begin{algorithmic}[1] 
    \REQUIRE ~~\\ 
    Universal unlabeled speech corpus $U$ sorted by length. \\
    Query speech corpus $Q$. \\
    A pre-trained Hubert model. \\
    Hyperparameters $N$, $C$ and $\lambda$.
    
    \ENSURE ~~\\ 
    Subset $S$ from $U$ and $|S|=C$
    
    \STATE Use the Hubert model to discrete the $U$ and $Q$.
    \STATE Calculate the $N$-gram model $P_{\widetilde{U}}(L)$ and $P_{\widetilde{Q}}(L)$,
    \STATE $P_{\widetilde{Q'}}(L) \gets (1- \lambda) P_{\widetilde{U}}(L) + \lambda P_{\widetilde{Q}}(L)$
    \STATE $S \gets \varnothing$, $r \gets |U|/C$
    \FOR{$i=0$ to $C-1$}
    \STATE Select the ($ir+1$)-th to the $(i+1)r$-th utterances in $U$ as $U_i$
    \STATE $u^i_{best}=\argmin_{u_j} \mathrm{SCD}(Q', S \cup \{u_j\}),$ where $u_j \in U_i$. 
    \STATE Add $u^i_{best}$ into $S$.
    \ENDFOR
    
    \RETURN  $S$. 
    \end{algorithmic}
\end{algorithm}

In practice, the available query corpus $Q$ is always small, it cannot fully represent the target scenario well. So directly using the $P_{\widetilde{Q}}(L)$ to calculate the SCD could make the $S$ overfit the $Q$. 
To increase the generalization ability of the selected set $S$, we use the interpolation method with $U$ and $Q$ as follows:
\begin{equation}
\begin{aligned}
P_{\widetilde{Q'}}(L) = \lambda P_{\widetilde{Q}}(L) + (1-\lambda)P_{\widetilde{U}}(L)
\end{aligned}
\end{equation}
\begin{equation}
\begin{aligned}
       S^* =  \argmin_S \mathrm{SCD}(Q',S), where \quad |S| = C 
\end{aligned}
\end{equation}

However, finding the global optimum solution $S^*$ is a NP-hard problem and the solution-space size is $|U| \choose C$. So we use the greedy-search method to find the local optimum solution to reduce the algorithm complexity. Details are shown in Algorithm \ref{alg:data-selection}.
As each utterance is visited only once, the complexity is $O(|U|K^N)$, $O(K^N)$ is the SCD complexity and $O(|U|)$ is the search complexity. And when $N$ is large, we can cut the rare grams to further reduce the complexity.

\begin{table}[!t]
    \small
    \caption{WER (\%) for models with different selected training corpus.}
    \centering
    \begin{threeparttable}
    \begin{tabular}{|l|l|c c|c c|c c|}
    \hline
    \multirow{2}{*}{Hours}&Selection& \multicolumn{2}{c|}{cmv} & \multicolumn{2}{c|}{ind} & \multicolumn{2}{c|}{aus} \\
    &Method & dev & test & dev & test & dev & test  \\
    \hline
    1 h &random &\textbf{30.1}&\textbf{31.8}&29.2&34.5&24.5&24.8\\
    &ind-label &32.0&34.2&\textbf{25.9}&32.5&32.8&32.7\\
    &ind-SCD  &31.0&32.5&27.5&\textbf{31.5}&27.4&28.0 \\
    &aus-label &35.5&38.4&38.9&39.6&\textbf{22.9}&\textbf{23.5}\\
    &aus-SCD &30.7&32.7&29.8&35.6&23.1&23.8\\
    \hline
    10 h &random &\textbf{22.8}&24.1&20.9&24.4&18.0&18.2\\
    &ind-label &24.6&25.7&\textbf{17.6}&23.0&25.7&26.3\\
    &ind-SCD &22.9&\textbf{23.9}&18.5&\textbf{21.4}&19.7&19.9 \\
    &aus-label &26.9&29.4&28.5&33.7&\textbf{14.9}&\textbf{15.6}\\
    &aus-SCD &23.3&24.6&21.7&25.0&16.7&17.3\\
    \hline
    100 h&random & 17.7&17.7&13.9&13.5&12.2&12.8\\
    &ind-SCD & \textbf{17.5}&\textbf{17.6}&\textbf{11.2}&\textbf{11.5}&12.2&13.6 \\
    &aus-SCD & 17.9&17.8&14.2&13.0&\textbf{11.3}&\textbf{11.9}\\
    \hline
    \end{tabular}
    \begin{tablenotes}
    \footnotesize
    \item The cmv stands for the official evaluation set. The ind and aus are our split accent evaluation set.
    \end{tablenotes}
    \end{threeparttable}
    \label{tab:result_main}
\end{table}

\section{Experiment}

\subsection{Dataset}
We conduct the experiments on the English Common Voice (CMV) v5.1 \cite{commonvoice} and take the accent as the basis of data selection. 
In CMV, volunteers worldwide contribute these audios by reading sentences provided by the website, and some of the volunteers also provide their accents.
We will select the training subset from the CMV and evaluate on the Indian (ind) and the Australia (aus) accents, which only account for 5\% and 4\% among the whole CMV.
For evaluation, besides the official evaluation set, we split a \textit{dev} set (5 hours) and a \textit{test} set (5 hours) for the ind-accent and aus-accent. These split parts will be excluded during data selection.

\subsection{Data selection}\label{sec:selection}

We use different data selection methods, including random selection, supervised accent-label selection, and the proposed unsupervised SCD selection to sample the training set. For the random selection, we shuffle the corpus and select the front utterances. For the accent-label selection, we only sample the speech with ind-label or aus-label. These two methods can be regarded as the lower bound and the upper bound of the experiments. 
For the SCD selection, we use the open source self-supervised Hubert-Large model \footnote{https://dl.fbaipublicfiles.com/hubert/hubert\_large\_ll60k.pt}\cite{hubert2} to discretize the speech into 500 cluster labels.
We count the distributions of the discrete corpus as a uni-gram ($N$=1). During the greedy-search, we use the \textit{dev-ind} or \textit{dev-aus} set as the $Q$ and $\lambda$ is adjusted from 0 to 1.
Finally, we fine-tune a Hubert as the downstream ASR model with  1 hour, 10 hours, and 100 hours of different selected training speech \footnote{We cannot sample the 100 hours set by the accent-label selection because the audios with the ind-label or aus-label are less than 100 hours.}. 



\begin{table}[!t]
    \small
    \caption{Influence of the discretization and the parameter $N$}
    \centering
    \begin{threeparttable}
    \begin{tabular}{|l|l|c c| c c|}
    \hline
    \multirow{2}{*}{Hours}&\multirow{2}{*}{Selection Method}&\multicolumn{2}{c|}{cmv} & \multicolumn{2}{c|}{ind} \\
    &&dev & test & dev & test  \\
    \hline
    1 h&SCD-MFCC-1-gram  & 31.1&32.7&27.2&33.5 \\
    &SCD-Hubert-B-1-gram & 31.4&33.2&27.9&32.4 \\
    &SCD-Hubert-L-1-gram & 31.0&\textbf{32.5}&27.5&\textbf{31.5} \\
    &SCD-Hubert-L-2-gram & \textbf{30.9}&32.8&27.3&32.4 \\
    &SCD-Hubert-L-3-gram &31.6&33.1&\textbf{27.1}&33.3 \\
    &CS-Hubert-L-1-gram &42.8&43.8&38.9&42.1 \\
    &CS-Hubert-L-2-gram &38.7&39.7&28.6&36.1 \\
    \hline
    10 h&SCD-MFCC-1-gram & \textbf{22.4}&23.7&18.5&24.1 \\
    &SCD-Hubert-B-1-gram & 22.7&23.7&18.6&22.0 \\
    &SCD-Hubert-L-1-gram & 22.8&23.9&18.5&21.4 \\
    &SCD-Hubert-L-2-gram & 22.5&\textbf{23.6}&17.0&21.3 \\
    &SCD-Hubert-L-3-gram & 22.7&24.0&16.8&21.9 \\
    &SCD-Text-Word-1-gram &24.1&25.8&23.0&27.1 \\
    &SCD-Text-Char-1-gram &25.4&28.3&24.9&31.1 \\
    &CS-Hubert-L-1-gram &27.8&28.2&18.2&21.6 \\
    &CS-Hubert-L-2-gram &26.3&27.3&\textbf{16.0}&\textbf{21.2} \\
    
    \hline
    
    \end{tabular}
    \begin{tablenotes}
    \footnotesize
    \item SCD-Text means that $P_{\widetilde{Q}}(L)$ and $P_{\widetilde{U}}(L)$ are calculated from the transcription.
    \item CS is short for contrastive sampling, which use $P_{\widetilde{Q}}(L)$ and $P_{\widetilde{U}}(L)$ to score and choose the utterances.
    \end{tablenotes}
    \end{threeparttable}
    \label{tab:result_feature_N}
\end{table}

\subsection{Main results}

Table \ref{tab:result_main} lists the main results of the experiments. According to the table, we can find that our proposed SCD selection can be better than the random selection on both ind-accent and aus-accent. Furthermore, compared to the supervised accent-label selection, the SCD selection can realize better results on the ind-accent and comparable results on the aus-accent. 
We can also find that the generalization of the SCD selection is better than the accent-label selection. The accent-label selection can benefit the recognition performance of the target accent but also hurt other accents at the same time. In contrast, the SCD selection can improve the target accent result with little influence on others. 
For example, on the 10 hours experiments, the ind-label selection can bring 5.7\% relative improvement on the \textit{test-ind} but also relatively increase 11.6\% and 44.5\% WER on the \textit{test-cmv} and \textit{test-aus}. 
In contrast, the ind-SCD selection can reduce the WER of \textit{test-ind} with 12.7\%, meanwhile, the WER increment on \textit{test-aus} is only 9.3\% and the \textit{test-cmv} result is even better.
By comparing the results with different training set sizes, we can find that the SCD selection can be more powerful with the sample size growing. 
When the selected training set increases from 1 to 100 hours, the relative improvement between SCD selection and random selection increases from 5.8\% to 14.8\% for ind-accent, and increases from 4.0\% to 7.0\% for aus-accent. Because the SCD selection is based on statistics, larger sample size will provide a more accurate probability distribution.

\subsection{Analysis}


\subsubsection{Influence of the discretization and N-gram}
We analysis the influence of the discretization and N-gram and show the result in Table \ref{tab:result_feature_N}. For the discretization, we also use the the K-means clustering with MFCC (the original target label of the Hubert) and the Hubert-Base to discrete the speech. 
We can find that all of them can exceed the random selection. This means that the SCD selection still can be useful even without a self-supervised model. 
Although similar performances are shown on the \textit{dev-ind}, the Hubert-Large discretization gets the best results on the \textit{test-ind}. This indicates that more informative discrete label can bring better generalization ability. 
For the N-gram, we also use larger $N$ to measure the distribution of the discrete label corpus and find that when $N$ becomes larger, the \textit{dev-ind} WER always decrease but the \textit{test-ind} result could become worse. This proves the selected training set over matches the \textit{dev-ind}. 
We believe the reason is that \textit{dev-ind} only contains 5 hours of speech, which is insufficient for a higher-order statistical model. 

\subsubsection{Compare with other methods}
We also use the transcription distribution and contrastive sampling for data selection like \cite{select-text} and \cite{google-ds}. And the results is also shown in Table \ref{tab:result_feature_N}. We can find that select data with the word or character is useless and even harmful in this task, because the transcription can not show the difference between different accents.
And the contrastive sampling can realize similar performance on the 10 hours \textit{ind-test} set, however, much worse on the \textit{cmv-test} set and the 1 hours training task. Because it tend to select similar utterances 
especially when the selected size is small.

\subsubsection{Influence of the interpolation factor $\lambda$}

\begin{figure}[!t]
\centering
\subfloat[1 hour results on  \textit{dev-ind}]{\includegraphics[width=0.47\columnwidth]{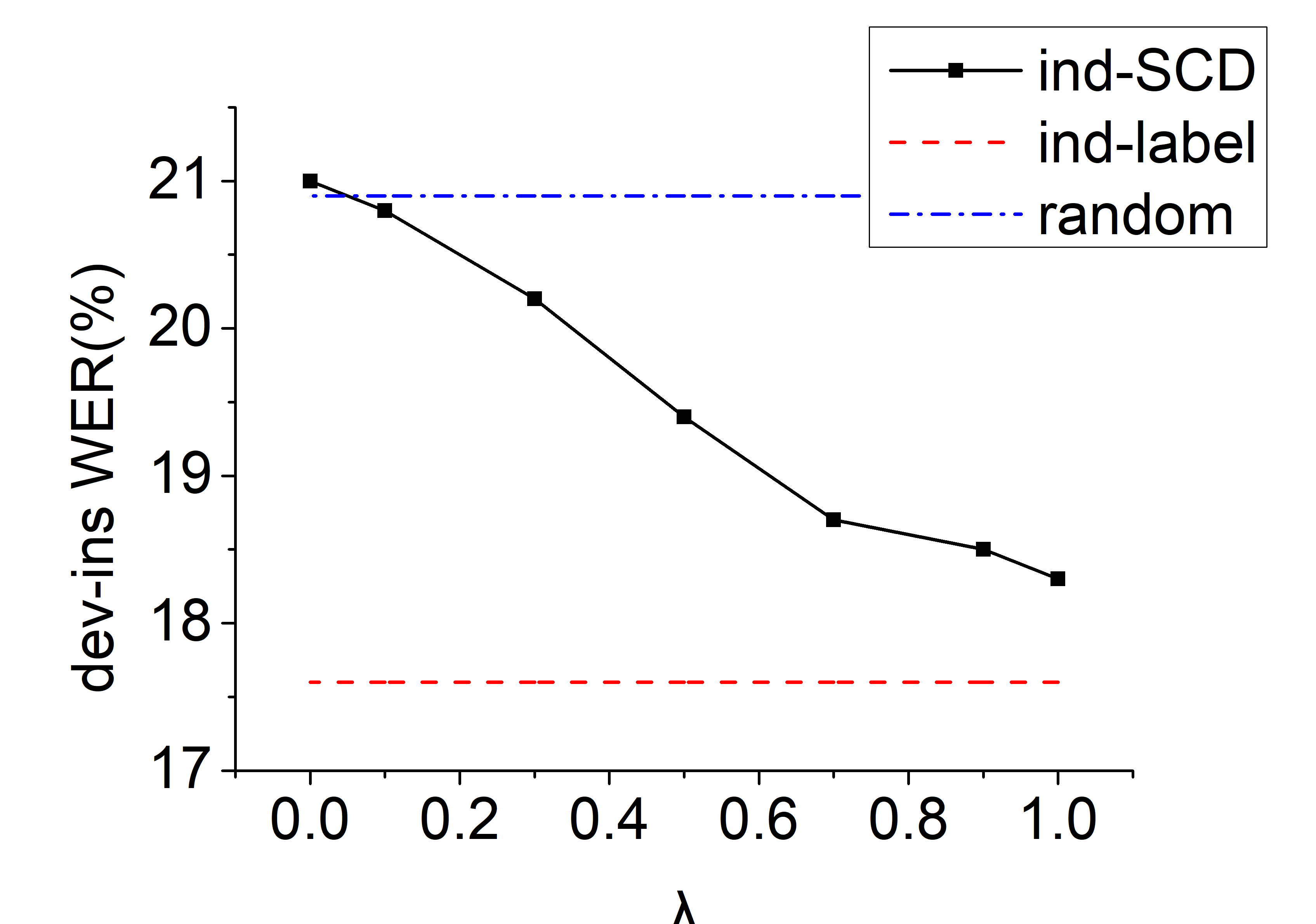}%
\label{fig_first_case}}
\hfil
\subfloat[1 hour results on  \textit{test-ind}]{\includegraphics[width=0.47\columnwidth]{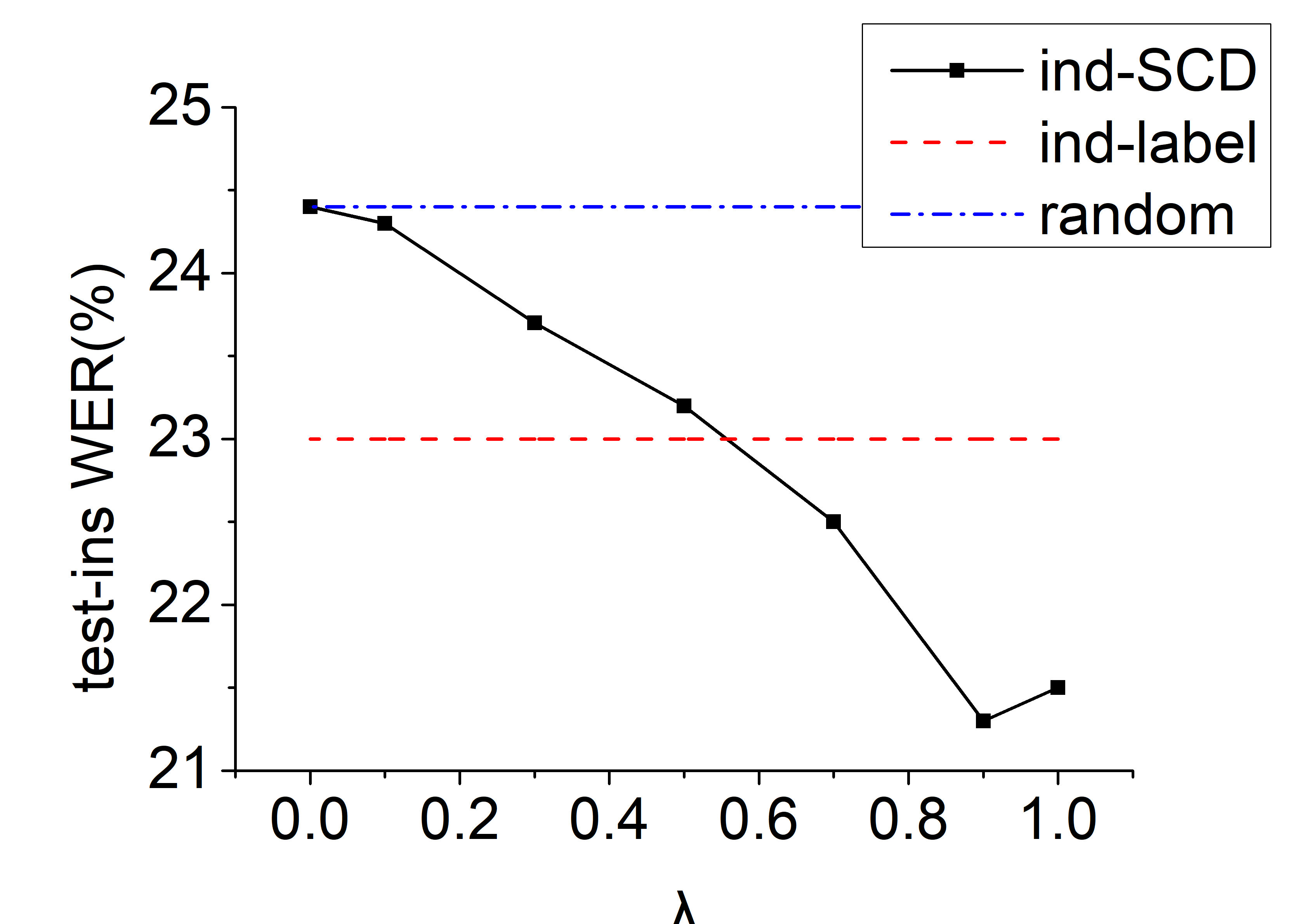}%
\label{fig_second_case}}
\hfil
\subfloat[10 hours results on  \textit{dev-ind}]{\includegraphics[width=0.47\columnwidth]{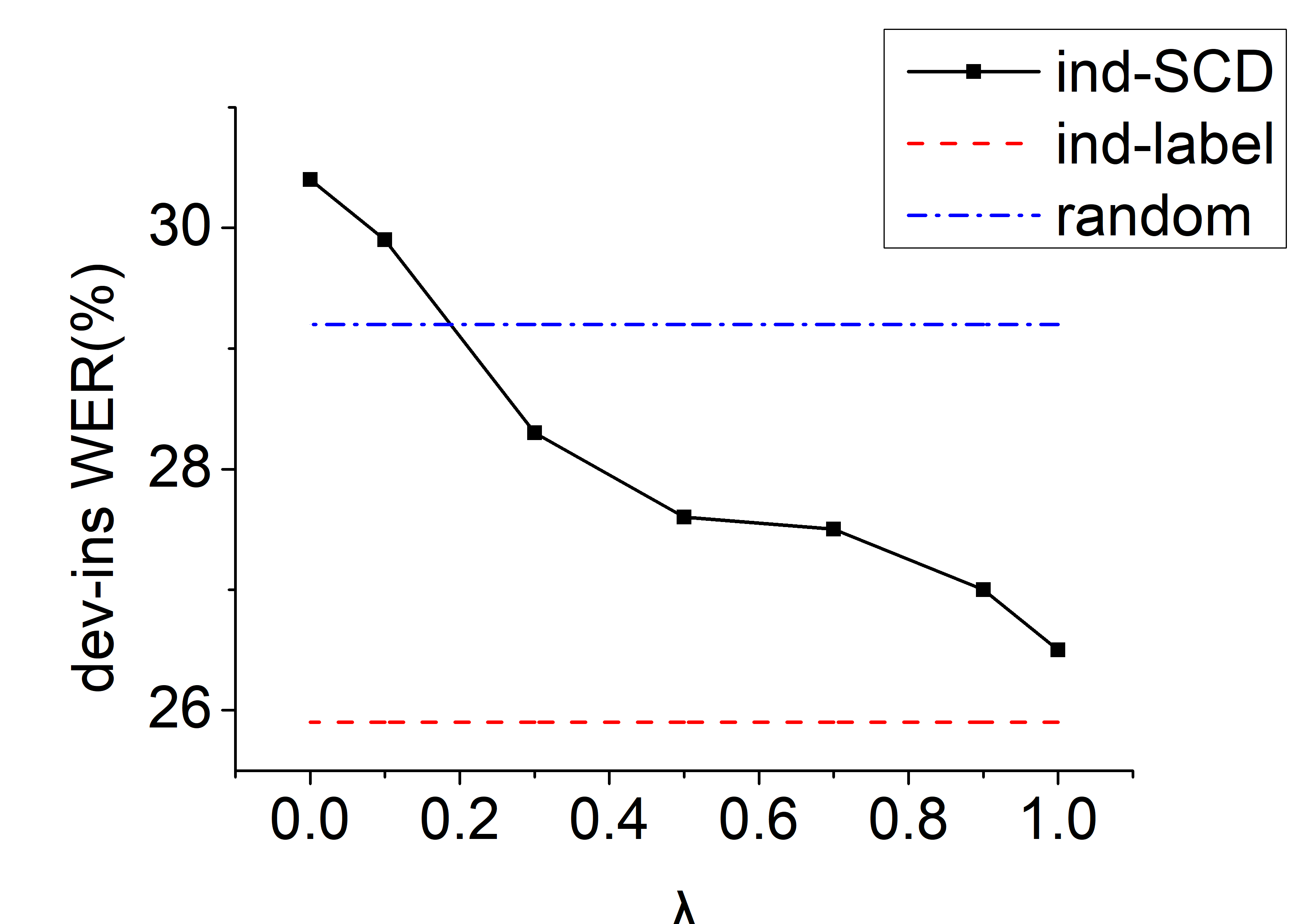}%
\label{fig_third_case}}
\hfil
\subfloat[10 hours results on  \textit{test-ind}]{\includegraphics[width=0.47\columnwidth]{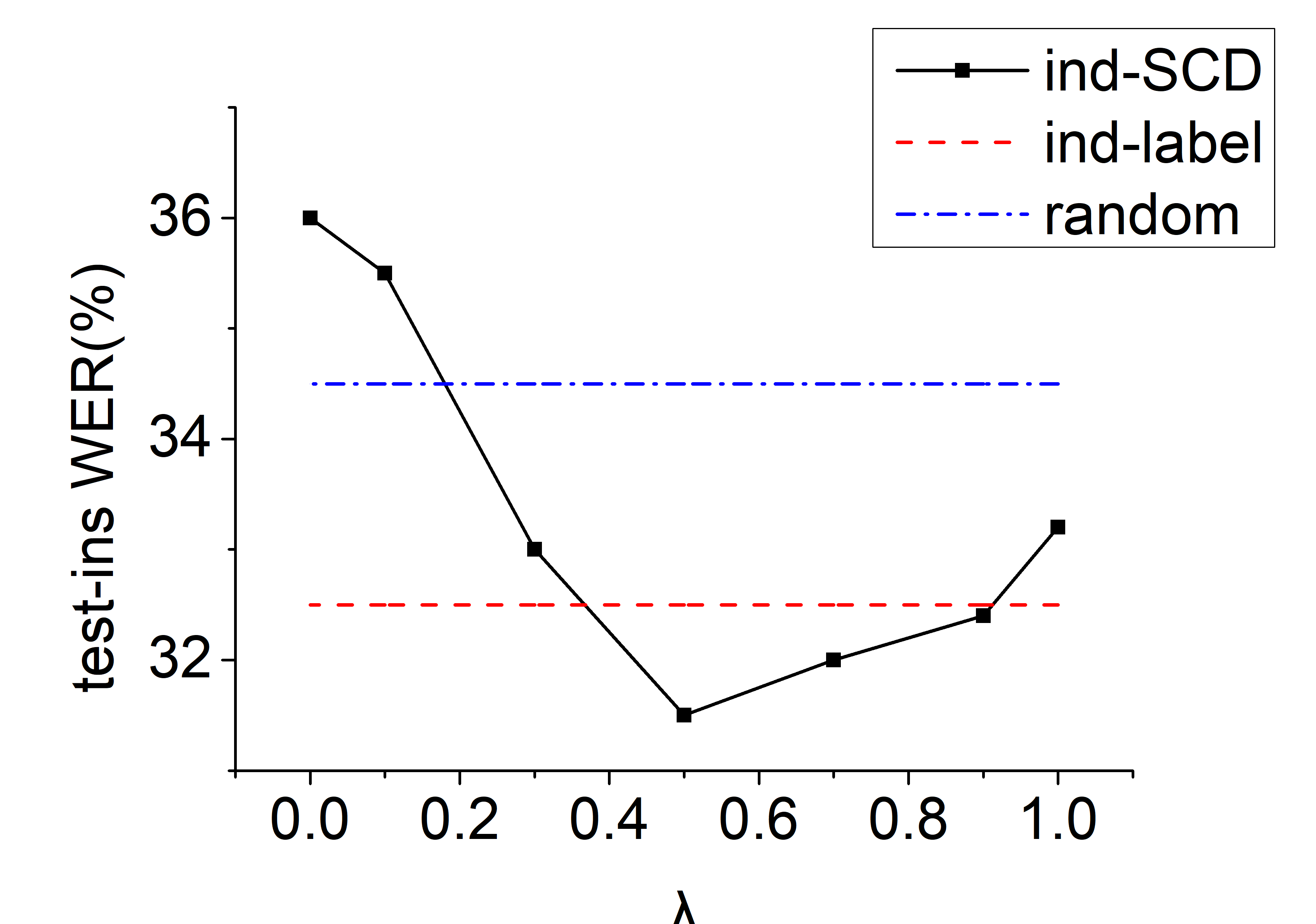}%
\label{fig_forth_case}}
\caption{Influence of the interpolation factor $\lambda$. We also draw the random selection (blue line) and ind-label selection (red line) in the figures as reference.}
\label{fig_lambda}
\end{figure}

We evaluate the influence of the interpolation factor $\lambda$, which is used to prevent the selected set from overfitting the query corpus. The experiments are based on the ind-accent with 1 or 10 hours training set, and the results are shown in Fig \ref{fig_lambda}. 
We can find that on the \textit{dev-ind}, the WER continuously decreases with the $\lambda$ growth, which means that the selected subset fits the \textit{dev-ind} better. 
However, for the \textit{test-ind}, the WER will reduce firstly and then rise again until $\lambda$ changes to 1.0. 
The phenomenon is more evident when the selected training set size is small.
This means that without interpolating with the universal corpus, the generalization ability of the SCD selection will be hurt.

\subsubsection{Selected training set construction}

\begin{figure}[!t]
\centering
\includegraphics[height=2in]{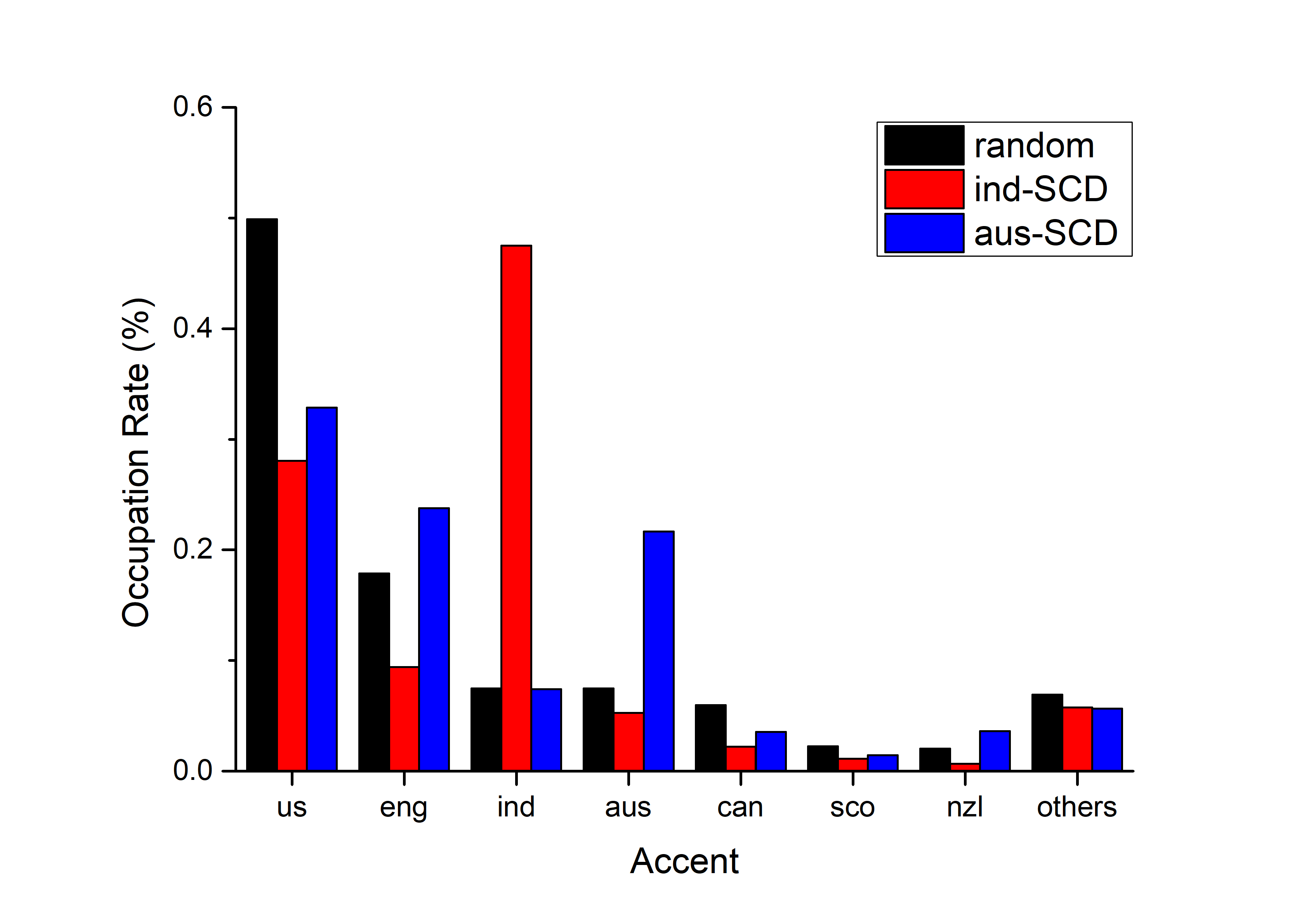}
\caption{Training set construction with different data selection method. The results are counted from the speech with accent label in the selected training set.}
\label{fig_construct}
\end{figure}

We draw the construction of three 10 hours training sets selected by random selection, ind-SCD selection, and aus-SCD selection in Fig \ref{fig_construct}.
We can find that the SCD selection set contains more target accented speech than the random selection. For example, when using random selection, the ind-accented and aus-accented speech only takes 7.5\%. 
For the ind-SCD and aus-SCD selection, the proportion of the ind-accented and aus-accented speech will increase to  48\% and 21\%. It should be noticed that during SCD selection, no accent label or transcription is used. 
Fig \ref{fig_construct} also shows the relationship between different accents. The ind-SCD selection choose more ind-accented speech rate and less other accent speech. However, the aus-SCD also samples more speech with England (eng) and Newzealand (nzl) labels beside the aus-accented speech. 
As we know, the aus-accent has closer relationship with the eng-accent and the nzl-accent than others.
This indicates that the proposed SCD is consistent with human cognition.

\section{Conclusion}
This study proposes SCD to measure the speech corpora similarity by the Hubert discrete label distribution and then select the training speech by the SCD. 
Compare to previous works, this SCD selection method can consider both acoustic and semantic information in the speech corpus and can also guarantee the diversity of the selected speech.
Experiments on different accents speech show that with the same training set size, the proposed SCD  selection can realize up to 14.8\% relative improvement than the random selection and also realize comparable even better performance than supervised selection with accent label.
As the SCD data selection method is independent from the transcription, we will apply this method on other audio tasks which need to sample the best training subset from a large-scale corpus.



%





\ifCLASSOPTIONcaptionsoff
  \newpage
\fi



\normalsize
\bibliographystyle{IEEEtran}
\bibliography{main}

\end{document}